\documentclass[conference]{IEEEtran}
\IEEEoverridecommandlockouts
\usepackage{cite}
\usepackage{amsmath,amssymb,amsfonts}
\usepackage{algorithmic}
\usepackage{graphicx}
\usepackage{amsmath}
\usepackage{amssymb}
\usepackage{mathtools}
\usepackage{textcomp}
\usepackage{xcolor}
\usepackage{caption}
\usepackage{hyperref}
\DeclarePairedDelimiter\abs{\lvert}{\rvert}%
\DeclarePairedDelimiter\norm{\lVert}{\rVert}%
\makeatletter
\let\oldabs\abs
\def\abs{\@ifstar{\oldabs}{\oldabs*}}
\let\oldnorm\norm
\def\norm{\@ifstar{\oldnorm}{\oldnorm*}}
\makeatother

\def\BibTeX{{\rm B\kern-.05em{\sc i\kern-.025em b}\kern-.08em
    T\kern-.1667em\lower.7ex\hbox{E}\kern-.125emX}}
\begin{document}

\title{Parsimonious Computing: A Minority Training Regime for Effective Prediction in Large Microarray Expression Data Sets}
\makeatletter
\newcommand{\linebreakand}{%
  \end{@IEEEauthorhalign}
  \hfill\mbox{}\par
  \mbox{}\hfill\begin{@IEEEauthorhalign}
}
\makeatother
\author{
  \IEEEauthorblockN{1\textsuperscript{st} Given Name Surname}
  \IEEEauthorblockA{\textit{dept. name of organization (of Aff.)} \\
    \textit{name of organization (of Aff.)}\\
    City, Country \\
    email address}
  \and
  \IEEEauthorblockN{2\textsuperscript{nd} Given Name Surname}
  \IEEEauthorblockA{\textit{dept. name of organization (of Aff.)} \\
    \textit{name of organization (of Aff.)}\\
    City, Country \\
    email address}
  \and
  \IEEEauthorblockN{3\textsuperscript{rd} Given Name Surname}
  \IEEEauthorblockA{\textit{dept. name of organization (of Aff.)} \\
    \textit{name of organization (of Aff.)}\\
    City, Country \\
    email address}
  \linebreakand 
  \IEEEauthorblockN{4\textsuperscript{th} Given Name Surname}
  \IEEEauthorblockA{\textit{dept. name of organization (of Aff.)} \\
    \textit{name of organization (of Aff.)}\\
    City, Country \\
    email address}
  \and
  \IEEEauthorblockN{5\textsuperscript{th} Given Name Surname}
  \IEEEauthorblockA{\textit{dept. name of organization (of Aff.)} \\
    \textit{name of organization (of Aff.)}\\
    City, Country \\
    email address}
}
\author{
 \IEEEauthorblockN{1\textsuperscript{st} Shailesh Sridhar}
\IEEEauthorblockA{Department of Computer Science\\
PES University\\
Bangalore, India \\
shailesh.sridhar@gmail.com}
  \and
  \IEEEauthorblockN{2\textsuperscript{nd} Snehanshu Saha}
\IEEEauthorblockA{Department of Computer Science and\\ Information Systems and APPCAIR \\
BITS Pilani K K Birla Goa Campus\\
Goa, India \\
snehanshu.saha@ieee.org}
  \and
  \IEEEauthorblockN{3\textsuperscript{rd} Azhar Shaikh}
\IEEEauthorblockA{Department of Electronics and\\ Communication\\
PES University\\
Bangalore, India \\
azhar199865@gmail.com}
  \linebreakand 
  \IEEEauthorblockN{4\textsuperscript{th} Rahul Yedida}
\IEEEauthorblockA{Department of \\Computer Science\\
North Carolina State University\\
Raleigh, USA \\
ryedida@ncsu.edu}
  \and
 \IEEEauthorblockN{5\textsuperscript{th} Sriparna Saha}
\IEEEauthorblockA{Department of Computer Science and\\ Engineering \\
IIT Patna\\
Patna, India \\
sriparna.saha@gmail.com }

}
\maketitle

\begin{abstract}
Rigorous mathematical investigation of learning rates used in back-propagation in shallow neural networks has become a necessity. This is because experimental evidence needs to be endorsed by a theoretical background. Such theory may be helpful in reducing the volume of experimental effort to accomplish desired results. We leveraged the functional property of Mean Square Error, which is Lipschitz continuous to compute learning rate in shallow neural networks. We claim that our approach reduces tuning efforts, especially when a significant corpus of data has to be handled. We achieve remarkable improvement in saving computational cost while surpassing prediction accuracy reported in literature. The learning rate, proposed here, is the inverse of the Lipschitz constant. The work results in a novel method for carrying out gene expression inference on large microarray data sets with a shallow architecture constrained by limited computing resources. A combination of random sub-sampling of the dataset, an adaptive Lipschitz constant inspired learning rate and a new activation function, A-ReLU  helped accomplish the results reported in the paper.
\end{abstract}


\section{Introduction}
{G}{ene} expression patterns are studied by microbiologists extensively to determine the genetic behaviour of cells. This is conventionally done via gene expression profiling, which is used to obtain and examine cell behaviour patterns in various scenarios such as drug treatment and disease.
The CMap (connectivity Map) project was launched with the intent of creating a reference collection of patterns \cite{lamb2006connectivity}.
There are approximately 22,000 genes across the entire human genome. However, it has been discovered that most of them are highly correlated. Accordingly, regulatory and target genes have been identified.
Researchers from the LINCS program, analyzing the CMap data found that around 80\% of the data can be captured using a set of ~1000 carefully selected genes. This inspired the development of the L1000 Luminex bead technology, which is able to measure the expression profiles of these  approximately 1000 genes, called landmark genes at a significantly lower cost \cite{peck2006method}. Using this data, the expression profiles of the other roughly 21,000 “target genes” can be inferred computationally. This computational inference is, however, challenging. Several techniques have been employed in the past for similar problems, such as Linear regression used by the LINCS researchers  \cite{hasty2001computational} and Kernel machines \cite{ye2013low}.

In recent years, due to the rapid growth of deep learning and neural networks, there have been several attempts to apply deep learning approaches to this problem. A recent significant work is D-GEX, a multi-task, multi-layer feedforward neural network, employing deep learning techniques such as dropout and momentum \cite{chen2016gene}. The authors compared the performance of this model to other machine learning methods such as k-NN regression and linear regression and show that deep learning achieves more accurate target gene predictions for gene expression. The best results of this model were obtained when an overwhelming 27,000 neurons were used across three hidden layers.

In contrast, we present a neural network that uses a substantially smaller number of neurons in a single hidden layer. It is reasonable to believe that the reduction in  computing units will limit the representational power of the network. In order to counter this limitation, we employ an adaptive learning rate based on the Lipschitz constant and an activation function(A-ReLU) \cite{Arelu} that is more suitable for the given task. Drawing inspiration from the D-GEX paper, we utilize the microarray expression data from the GEO \cite{edgar2002gene} project.

\subsection{Motivation \& Contribution}
A gene expression profile consists of thousands of genes. Even a set of landmark genes consists of several hundreds. While deep learning architectures with several thousands of neurons are able to satisfactorily solve such problems, they are computationally intensive and prohibitively expensive. The authors of D-GEX \cite{chen2016gene} used an NVIDIA GTX TITAN Z Graphics card with dual GPUs to train. Such computing infrastructure is scarce and orthogonal to the philosophy of Parsimonious Computing, propagated in our work. Our goal is to show that a much smaller neural architecture can be used for the same task without suffering from unacceptable prediction error and can be trained with much smaller computing infrastructure requirements. The following theoretical contributions help in solving the gene inference problem with minimal computing infrastructure, called Parsimonious Computing model proposed in our work.

a) \textbf{Lipschitz constant adaptive learning rate} \\
The magnitude of the learning rate plays a huge role in determining the time a neural network will take to converge. Too small a learning rate results in a very small rate of convergence and possible premature convergence at a local minima, while a learning rate which is too large may not be able to converge at all. To counter this,  adaptive learning rate schedulers have been proposed, so that the learning rate is adequately large when away from the global minima, and appropriately small when the network is close to it. Here we propose a learning rate scheduler which is calculated from the following formula:

\begin{equation}
\boxed{
	\max\limits_{i,j} \norm{ \frac{\partial E}{\partial w^{[L]}_{ij}} } = \frac{1}{m} \left( K_a + \norm{\textbf{y}} \right) K_z
}
\label{eq:reg:nn}
\end{equation}

where,
\begin{itemize}
   \item $K_a$ denotes the maximum of the activation in the final layer of the network 
   \item $K_z$ denotes the maximum of the activation in the penultimate layer
   \item $y$ denotes the target values 
   \item $m$ is the number of training examples
\end{itemize}

b) \textbf{A-ReLU as an activation function} \\
The original D-GEX network utilizes tanh as an activation function. Though quite effective, activation functions such as tanh and sigmoid suffer from the vanishing gradient problem as the gradient is squashed between a small range of values. ReLU overcomes this issue by providing a constant gradient value for all inputs. However, ReLU has issues of its own, as it is non-differential at 0.
In this work, we use A-ReLU as an activation function, a relatively new activation function defined by the equation

\[ \begin{cases} 
     kx^{n} &x\geq  0\\
      0 & x <0 \\
   \end{cases}
\]

We show experimentally that it performs better than other classical activation functions in this regression task. We claim to have established the following major items.

\begin{itemize}
    \item Detailed proof of Lipschitz Adaptive Learning Rate (LALR) for Mean Square Loss (MSE) in training a shallow neural network.
    \item A novel activation function, A-ReLU to demonstrate better performance compared to established activation units
    \item Propose Parsimonious Computing, a philosophy that advocates practice of frugality in computing resources, by leveraging deep mathematical insights; in particular, a combination of the above two items.
    \item Provide empirical evidence that the effect of Lipschitz learning is equivalent to dropping one hidden layer of $9000$ neurons in the deep learning architecture used in the base paper\cite{chen2016gene}, while maintaining comparable prediction accuracy. In other words, a shallow network of 2 hidden layers of 9000 neurons, powered by Parsimonious Computing model matches the performance of a deep neural net with 3 hidden layers of $9000$ neurons, twice as large as our shallow neural network. Table I provides evidence in support of this claim.
\end{itemize}

\section{Literature Survey}
The field of micro-array bioinformatics has witnessed tremendous growth with a large focus on gene expression profiling.  Gene expression profiling  was used to identify bio-markers for Parkinson's disease in \cite{diao2012gene}.   Molecular information obtained from Micro-Array data is leveraged to predict if a tumour is cancerous or not in a supervised setting. The same information is used to discover new types of tumours in an unsupervised setting in \cite{perez2007microarrays}.  A comprehensive comparison of applying different machine learning techniques like SVM, RBF multi-layer perceptron, Random Forest etc. and feature selection methods like chi-squared and CSF for classification on micro-array gene expression data was given in \cite{pirooznia2008comparative}.  Correlation between gene profiles is exploited by gene regulatory networks\cite{bansal2007infer} to identify landmark and target genes. This paved the way for research in building computational inference models to predict target gene profiles using landmark genes. A computational model based on linear regression was proposed by \cite{hasty2001computational}, but was not complex enough to capture intrinsic non-linearity present within micro-array data. The number of independent regulatory modules between the landmark and target genes are inferred from the rank of their connectivity matrix and then is used as a low rank regularisation constraint to formulate gene expression inference as a multi-target linear regression problem in \cite{ye2013low}. Recent emergence of deep learning methods has led to its application as a diagnostic tool for gene expression analysis. DeepCC\cite{gao2019deepcc}, a deep learning framework for cancer molecular sub-type classification, outperforms all traditional machine learning methods on 13 independent data sets based on Affymetrix platforms.  A 3 layer feed-forward neural network named D-GEX is used to predict gene expression values of target genes in \cite{chen2016gene}. This method outperforms all previous methods by a large margin and is used as a baseline for our work.

A fundamental problem faced by neural networks is that of finding optimal values for its learning rate. Recent works agree that a non-monotonic learning rate scheduling system would offer faster convergence \cite{smith2017cyclical},\cite{smith2019super}. Of late, there has been some development in finding novel ways to adaptively change the learning rate of a neural network. The results have  theoretical, intuitive and empirical support and rely on non-monotonic scheduling of the learning rate. The method we use here, also yields a non-monotonic learning rate, but does not follow any predefined shape. Contrary to the trend of using deeper neural networks for regressing gene expression values, we exploit wider shallow neural networks. We claim these are as effective as their deep counterparts with the added benefit of cheap training due to the reduction in number of parameters.

The remainder of the paper is organized in the following manner: Section III describes data and and its handling by constrained computing resources. This is followed by Parsimonious Computing in section IV, the motivation behind tackling such a corpus of data in a cost effective way.  Section IV, thus, describes the methods which help accomplish Parsimonious Computing goals, in other words make cheap  training a feasible option. Section V discusses experimental settings and interventions, followed by detailed results in section VI.

\section{Datasets}
GEO (Gene Expression Omnibus) is a public functional genomics data repository from which The Broad Institute produced the GEO expression data. The data set consists of 129,158 gene expression profiles or samples obtained from the Affymetrix microarray platform. Each profile is associated with 22,268 gene probes, Out of which 978 are landmark genes and the remaining 21,290 are target genes whose values are to be predicted. Authors in the base paper\cite{chen2016gene} processed the data into a simpler and easier to handle format, by implementing first quantile normalization of the data into a numerical range between 4 and 15, followed by removing any duplicates. The final, ready-to-use data has 111,009 profiles.

The GEO dataset is enormous, with just the training data consuming more than 3GB of space. Given that all training was carried out using a single GPU with 4GB ram, we used random subsampling along with the 50:50 split employed by the D-GEX authors\cite{chen2016gene}.

In order to represent the entire GEO-tr dataset appropriately while ensuring that the computing infrastructure utilized can handle the scale of data used, the network is trained on six random sub-samples of the data set, each containing 20,000 data points or slightly less than 1/5th of it. A representative MAE was obtained by averaging the MAE obtained across each individual subsample. This average  is considered so that any disparity among sub-samples could be taken into account.

\section{Parsimonious Computing: Our contribution in Cheap Training}

In this section we discuss the methods used in our model that enable us to achieve results very close to those reported by the D-GEX method \cite{chen2016gene} but without requiring the same amount of compute power. All experiments were carried out on a laptop with an Intel i5 processor and an NVIDIA GTX-1050Ti graphics card. The GPU utilized has 4GB RAM associated with it.

\subsection{Lipschitz Adaptive Learning Rate}

Recently, there has been a lot of work on finding novel ways to adaptively change the learning rate. These have both theoretical \cite{seong2018towards} and intuitive, empirical \cite{smith2019super, smith2017cyclical} backing. These works rely on non-monotonic scheduling of the learning rate. Authors in \cite{smith2017cyclical} argue for cyclical learning rates. Our proposed method also yields a non-monotonic learning rate, but does not follow any predefined shape.
we propose a novel theoretical framework to compute large, adaptive learning rates for use in gradient-based optimization algorithms. We start with a presentation of the theoretical framework and the motivation behind it, and then derive the mathematical formulas to compute the learning rate on each epoch.  

Our results show that, compared to standard choices of learning rates, our approach converges quicker and achieves better results. Our approach exploits functional properties of the loss function, and only makes two minimal assumptions on the loss function: it must be Lipschitz continuous\cite{lipschitz} and (at least) once differentiable. Commonly used loss functions satisfy both these properties.  We argue that the use of Lipschitz constants to determine learning rate greatly improves convergence in comparison with standard learning rate choices. We present empirical evidence of our claims in the results section. This is a departure from the approach of manually tuning learning rates.
\subsubsection{Theoretical Framework}
\label{sec:theory}

For a function, the Lipschitz constant is the least positive constant $L$ such that 

\begin{equation}
    \left\Vert f(\textbf{w}_1) - f(\textbf{w}_2)\right\Vert \leq L \left\Vert \textbf{w}_1 - \textbf{w}_2 \right\Vert
\end{equation}

for all $\textbf{w}_1$, $\textbf{w}_2$ in the domain of $f$. From the mean-value theorem for scalar fields, for any $\textbf{w}_1, \textbf{w}_2$, there exists $\textbf{v}$ such that 

\[
    \begin{aligned}
        \norm{f(\textbf{w}_1) - f(\textbf{w}_2)} &= \norm{ \nabla_{\textbf{w}} f(\textbf{v})} \norm{ \textbf{w}_1-\textbf{w}_2} \\
        &\leq \sup\limits_{\textbf{v}} \norm{\nabla_{\textbf{w}} f(\textbf{v})} \norm{\textbf{w}_1-\textbf{w}_2}
    \end{aligned}
\]
Thus, $\sup\limits_{\textbf{v}} \norm{\nabla_{\textbf{w}} f(\textbf{v})}$ is such an $L$. Since $L$ is the least such constant, 

\begin{equation}
    L \leq \sup\limits_{\textbf{v}} \norm{ \nabla_{\textbf{w}} f(\textbf{v}) }
\end{equation}
In this paper, we use $\max \norm{\nabla_{\textbf{w}} f}$ to derive the Lipschitz constants. Our approach makes the minimal assumption that the functions are Lipschitz continuous and differentiable up to first order only \footnote{Note this is a weaker condition than assuming the gradient of the function being Lipschitz continuous. We exploit merely the boundedness of the gradient.}. Because the gradient of these loss functions is used in gradient descent, these conditions are guaranteed to be satisfied. 

By setting $\alpha = \frac{1}{L}$, we have $\Delta \textbf{w} \leq 1$, constraining the change in the weights. We stress here that we are not computing the Lipschitz constants of the \textit{gradients} of the loss functions, but of the losses themselves. Therefore, our approach merely assumes the loss is $L$-Lipschitz, and not $\beta$-smooth. We argue that the boundedness of the effective weight changes makes it optimal to set the learning rate to the reciprocal of the Lipschitz constant. This claim, while rather bold, is supported by our experimental results.
\subsubsection{Significance of Lipschitz constant (LC)}
The Lipschitz constant (LC) has found a variety of uses in computing and applications. The central condition to the existence and uniqueness of solutions to first order system of differential equations of the form $y'(t)=f(t,y(t)$ is LC of $f$. The existence of LC guarantees contraction and eventually a fixed point i.e. solution to the above system\cite{banach1922operations} and saves the trouble of computing an analytical solution to the system above. Finding an LC is equivalent to to the fact that the function, $f$ possesses Lipschitz continuity. Given, $f:R\xrightarrow{}R, $ there exists an $L$ such that
\begin{equation*}
    \left\Vert f(\textbf{x}) - f(\textbf{y})\right\Vert \leq L \left\Vert \textbf{x} - \textbf{y} \right\Vert
\end{equation*}
Consequently, \begin{equation*}
   \frac{\left\Vert f(\textbf{x}) - f(\textbf{y})\right\Vert}{\left\Vert \textbf{x} - \textbf{y} \right\Vert}  \leq L \end{equation*} 
This is equivalent to stating that computing a LC of a function (loss function, in our case) is identical to computing the maximum of the derivative of $f$, via the Mean Value theorem. For loss functions which are differentiable, we can easily compute LCs and therefore find the bound on the derivatives to be used in deep neural network training. Mean Square Loss (MSE) satisfy the conditions of differentiability and hence Lipschitz continuity. We compute the LC of MSE enabling us to arrive at adaptive learning rate formulation, Lipschitz Adaptive Learning rate (LALR).
\subsubsection{Deriving the Lipschitz constant for neural networks}

For a neural network that uses the sigmoid,  (or A-ReLU), or softmax activations, it is easily shown that the gradients get smaller towards the earlier layers in backpropagation. Because of this, the gradients at the last layer are the maximum among all the gradients computed during backpropagation. If $w^{[l]}_{ij}$ is the weight from node $i$ to node $j$ at layer $l$, and if $L$ is the number of layers, then

\begin{equation}
	\max\limits_{h, k} \norm{\frac{\partial E}{\partial w^{[L]}_{hk}}} \geq \norm{\frac{\partial E}{\partial w^{[l]}_{ij}}} \forall\  l, i, j \label{eq:int:1}
\end{equation}
\subsubsection{Least-squares cost function}
\label{sec:leastsq}
For the least squares cost function, we will compute the Lipschitz constant for linear regression where the output is continuous. We will then prove the equivalence of the general result with regression in neural networks and derive the former as a special case of the latter.

\subsubsection{Linear regression}
We have,
\[
    g(\textbf{w}) = \frac{1}{2m}\sum\limits_{i=1}^m \left(\textbf{x}^{(i)} \textbf{w} - y^{(i)}\right)^2
\]
Thus,
\[
    \begin{aligned}
        g(\textbf{w}) - g(\textbf{v}) &= \frac{1}{2m}\sum\limits_{i=1}^m \left(\textbf{x}^{(i)} \textbf{w} - y^{(i)}\right)^2 - \left(\textbf{x}^{(i)} \textbf{v} - y^{(i)}\right)^2 \\
        &= \frac{1}{2m}\sum\limits_{i=1}^m \left( \textbf{x}^{(i)}(\textbf{w}+\textbf{v}) - 2y^{(i)}\right) \left( \textbf{x}^{(i)} (\textbf{w}-\textbf{v}) \right) \\
        &= \frac{1}{2m}\sum\limits_{i=1}^m \left( (\textbf{w}+\textbf{v})^T \textbf{x}^{(i)T} - 2y^{(i)}\right) \left( \textbf{x}^{(i)} (\textbf{w}-\textbf{v}) \right) \\
        &= \frac{1}{2m}\sum\limits_{i=1}^m \left( (\textbf{w} + \textbf{v})^T \textbf{x}^{(i)T}\textbf{x}^{(i)} - 2y^{(i)}\textbf{x}^{(i)} \right) (\textbf{w}-\textbf{v}) 
    \end{aligned}
\]
The penultimate step is obtained by observing that $(\textbf{w}+\textbf{v})^T \textbf{x}^{(i)T}$ is a real number, whose transpose is itself.

At this point, we take the norm on both sides, and then assume that $\textbf{w}$ and $\textbf{v}$ are bounded such that $\left\Vert \textbf{w} \right\Vert, \left\Vert \textbf{v} \right\Vert \leq K$. Taking norm on both sides,
\[
    \boxed{
        \frac{\left\Vert g(\textbf{w}) - g(\textbf{v}) \right\Vert}{\left\Vert \textbf{w} - \textbf{v} \right\Vert} \leq \frac{K}{m}\left\Vert \textbf{X}^T\textbf{X} \right\Vert + \frac{1}{m} \left\Vert\textbf{y}^T \textbf{X} \right\Vert
    }
\]
We are forced to use separate norms because the matrix subtraction $2K \textbf{X}^T\textbf{X} - 2\textbf{y}^T \textbf{X}$ cannot be performed. The RHS here is the Lipschitz constant. Note that the Lipschitz constant changes if the cost function is considered with a factor other than $\frac{1}{2m}$.
\subsubsection{Regression with neural networks}
Let the loss be given by
\begin{equation}
	E(\textbf{a}^{[L]}) = \frac{1}{2m} \left( \textbf{a}^{[L]} - \textbf{y} \right)^2 \label{eq:reg:1}
\end{equation}
where the vectors contain the values for each training example. Then we have,
\begin{align*}
	E(\textbf{b}^{[L]}) - E(\textbf{a}^{[L]}) &= \frac{1}{2m} \left( \left( \textbf{b}^{[L]} - \textbf{y} \right)^2 - \left( \textbf{a}^{[L]} - \textbf{y} \right)^2 \right) \\
	&= \frac{1}{2m} \left( \textbf{b}^{[L]} + \textbf{a}^{[L]} - 2\textbf{y} \right) \left( \textbf{b}^{[L]} - \textbf{a}^{[L]} \right)
\end{align*}
This gives us,
\begin{align}
	\frac{\lVert E(\textbf{b}^{[L]}) - E(\textbf{a}^{[L]}) \rVert}{\lVert \textbf{b}^{[L]} - \textbf{a}^{[L]} \rVert} &= \frac{1}{2m} \lVert \textbf{b}^{[L]} + \textbf{a}^{[L]} - 2\textbf{y} \rVert \nonumber \\
	& \leq \frac{1}{m} \left( K_a + \norm{\textbf{y}} \right) \label{eq:reg:2}
\end{align}
where $K_a$ is the upper bound of $\norm{\textbf{a}}$ and $\norm{\textbf{b}}$. A reasonable choice of norm is the 2-norm.

By equation (13) (please see the subsection below, Equivalence of the constants), the second term on the right side of the equation is the derivative of the activation with respect to its parameter. Notice that if the activation is sigmoid or softmax, then it is necessarily less than 1; if it is ReLU type, it is either 0 or 1. Therefore, to find the maximum, we assume that the network is comprised solely of ReLU type activations, and the maximum of this is 1.

From (13) and 'equivalence of constants' calculations, we obtain
\begin{equation}
\boxed{
	\max\limits_{i,j} \norm{ \frac{\partial E}{\partial w^{[L]}_{ij}} } = \frac{1}{m} \left( K_a + \norm{\textbf{y}} \right) K_z
}
\label{eq:reg:nn}
\end{equation}
The Learning Rate to be used is hence a reciprocal of this calculated value. For example, in one of our experiments, after initialization, the values of the constants turned out to be the following:\\
$K_z$ = 983.88; $K_a$ = 142.86 \& $y = 4329.24$. Substituting these values in the equation and scaling down by multiplying by a factor of 0.3, we arrive at a learning rate of  approximately $1.36 \times 10^{-4}$. The scale-down factor is intuitive and considered "on-the-fly" to mitigate a likely exploding gradient problem by A-ReLU.
\subsubsection{Equivalence of the constants}
The equivalence of the above two formulas is easy to see by understanding the terms of \eqref{eq:reg:nn}. Let us define 
\begin{equation}
    K_z = \max\limits_j \norm{a_j^{[L-1]}}
\end{equation}
In any layer, we have the computations
\begin{align}
	z^{[l]} &= W^{[l]T}a^{[l-1]} + b^{[l]} \label{eq:int:2} \\
	a^{[l]} &= g(z^{[l]}) \label{eq:int:3} \\
	a^{[0]} &= X \label{eq:int:4}
\end{align}
Thus, the gradient with respect to any weight in the last layer is computed via the chain rule as follows.
\begin{align}
	\frac{\partial E}{\partial w^{[L]}_{ij}} &= \frac{\partial E}{\partial a^{[L]}_j}\cdot \frac{\partial a^{[L]}_j}{\partial z^{[L]}_j}\cdot \frac{\partial z^{[L]}_j}{\partial w^{[L]}_{ij}} \nonumber  \\
	&= \frac{\partial E}{\partial a^{[L]}_j}\cdot \frac{\partial a^{[L]}_j}{\partial z^{[L]}_j}\cdot a^{[L-1]}_i \label{eq:int:5}
\end{align}
This gives us
\begin{equation}
	\max\limits_{i,j} \abs{\frac{\partial E}{\partial w^{[L]}_{ij}}} \leq \max\limits_j \abs{ \frac{\partial E}{\partial a^{[L]}_j} }\cdot \max\limits_j \abs{ \frac{\partial a^{[L]}_j}{\partial z^{[L]}_j} }\cdot \max\limits_j \abs{ a^{[L-1]}_j } \label{eq:int:6}
\end{equation}
Because a linear regression model can be thought of as a neural network with no hidden layers and a linear activation, and from \eqref{eq:int:4}, we have, 
\[
    \textbf{a}^{[L-1]} = \textbf{a}^0 = \textbf{X}
\]
and therefore 
\begin{equation}
    K_z = \max\limits_j \norm{a_j^{[L-1]}} = \norm{\textbf{X}}
    \label{eq:reg:Kz}
\end{equation}
Next, observe that $K_a$ is the upper bound of the final layer activations. For a linear regression model, we have the ``activations" as the outputs: $\hat{\textbf{y}} = \textbf{W}^T \textbf{X}$. Using the assumption that $\norm{\textbf{W}}$ has an upper bound $K$, we obtain
\begin{equation}
    K_a = \max\limits \norm{\textbf{a}^{[L]}} = \max \norm{\textbf{W}^T \textbf{X}} = \max \norm{\textbf{W}} \cdot \norm{\textbf{X}} = K\norm{\textbf{X}}
    \label{eq:reg:Ka}
\end{equation}
Substituting \eqref{eq:reg:Kz} and \eqref{eq:reg:Ka} in \eqref{eq:reg:nn}, we obtain
\begin{align*}
    \max\limits_{i,j} \norm{ \frac{\partial E}{\partial w^{[L]}_{ij}} } &= \frac{1}{m} \left( K_a + \norm{\textbf{y}} \right) K_z \\
    &= \frac{1}{m}\left( K \norm{\textbf{X}} + \norm{\textbf{y}} \right) \norm{\textbf{X}} \\
    &= \frac{K}{m}\norm{\textbf{X}^T\textbf{X}} + \frac{1}{m}\norm{\textbf{y}^T \textbf{X}}
\end{align*}

\subsection{A-ReLU}
We have employed the use of the activation function A-ReLU\cite{Arelu} in our experiments. A-ReLU is a continuous and differentiable approximation of the ReLU Activation function, proven to be differentiable at $0$ too, thereby alleviating the problem of undefined gradients in the neighbourhood of $0$ faced by ReLU. Moreover, it is straightforward to show that
\begin{itemize}
    \item A-ReLU does not suffer from local minima problems
    \item A-ReLU does not admit of saddle points
    \item A-ReLU admits of a fixed point (easy to show by virtue of Intermediate Value Theorem) thereby ensuring optima.
    \item The exploding gradient is easy to control
\end{itemize}
What needs to be considered is the lack of tuning efforts activation functions usually need. This was accomplished using approximation techniques and Hausdorff distance. ReLU is an offshoot of SBAF \cite{Arelu}. Let us consider the activation function, SBAF:
\begin{equation}
y = \frac{1}{1+kx^\alpha(1-x)^{1-\alpha}}
\end{equation}
$\alpha$ + $\beta$ =1 where $\alpha$ $>$ 0 and  $\beta >0$. We show the $k, \alpha$ values for which SBAF approximates to A-ReLU activation function. $k=1, \alpha= 1$; SBAF becomes $\frac{1}{1+x}$ which upon binomial expansion (restricting to first order expansion assuming $0<x<1$) yields $y= 1-x= 1-\text{ReLU}$.  Approximate ReLu (A-ReLu) is motivated by the fact it is a least square approximation of ReLu such that AReLu is continuous and differentiable at $x=0$ unlike ReLu and could also be derived from the generic family of activation functions detailed in [6]. Therefore, least square optimization is the way forward to determine the optimal parameters of A-ReLU. This is presented below.\\
Consider the function $f(x) = kx^n$.
We know that the ReLU activation function is $y = \max(0,x)$. We need to approximate the values $n$ and $k$ such that $f(x)$ approximates to the ReLU activation function over a fixed interval.  Define, $
\text{Relative error} = \frac{||f(x) - y||}{||y||}$. Let the minimum tolerable error be $\epsilon < 10^{-3}$.Thus, assuming a error threshold, \begin{align*}
    \frac{||f(x) - y||}{||y||} &<=\epsilon<10^{-3}\\
    ||f(x) - y||&\leq 10^{-3}||y||\\
    ||f(x)||&\leq ||y||(10^{-3} +1)\\
    ||f(x)||&\leq 1.001||y||
\end{align*}
Since $f(x) = kx^n$ approximates the positive half (i.e., when $x>0$) of the ReLU activation function, $y = max(x,0)$, the value of y when $x>0$ can be written as: $||y|| = ||x||$.
Using this value in the error calculation, we rewrite the error approximation as, 
\begin{align*}
    ||kx^n||&\leq 1.001||x||\\
    ||kx^{n-1}||&\leq 1.001
\end{align*}
The above is an optimization problem i.e. $\min \left\Vert kx^{n-1}\right\Vert$ subject to the constraints $k>0, n>1, -10<x<10$. We obtain the following bounds on $k$ and $n$:
\begin{align*}
    0<k<1; 
    1<n<2
\end{align*}
Therefore, we obtain the following continuous approximation of ReLU:
\[ \begin{cases} 
     kx^{n} &x\geq  0\\
      0 & x <0 \\
   \end{cases}
\]
where $0<k<1, 1<n<2, -10<x<10$. More precisely, the approximation to the order of $10^{-3}$ is $k=0.54, n=1.3$. We used the parameter values $k,\alpha$ in the ballpark range while training the network. This ballpark range agrees with the Mathematical notion of $\epsilon-$ neighborhood of the theoretically computed values.

\section{Experiments And Results}
The variables in our experiments are the type of activation function used, the number of hidden layer neurons, the number of epochs and the type of learning rate policy used. For each experiment, we choose one activation function from a pool of four (Sigmoid, Tanh, ReLU, A-ReLU), either 500 or 1500 or 3000 hidden neurons, one among fixed learning rate policy, the Decay Factor based Scheduler employed in \cite{chen2016gene} and the Lipschitz learning rate policy for 200 epochs. In \cite{chen2016gene}, the authors used a pair of bounds on the higher and lower values of the learning rate ($5 \times 10^{-4}$) and ($1 \times 10^{-5}$). We use the same pair of bounds in our training. We only use a single hidden layer in most experiments as opposed to upto 3 (with upto 9000 hidden neurons in each layer) used in the D-GEX(D) architecture\cite{chen2016gene} which uses a decay factor based scheduler to vary the learning rate, and report a MAE which is very close to the one reported by the base paper authors. Hence our proposed model can effectively perform as well as the deep model but with a much smaller number of parameters.Due to the shallowness of the model used, it can be trained on low compute hardware, as shown in our experiments. We train our model for 100 iterations every epoch with a batch size of 200 on a single Nvidia 1050Ti GPU. From here onwards we will refer to the models using a Decay based LR(Base paper), Lipschitz Adaptive LR(Our proposed model) and fixed LR(for comparison) as D-GEX(D), D-GEX(L) and D-GEX(F). All experiments carried out for the aforementioned models and results obtained are without dropout.

We trained our modified D-GEX(L) architecture on GEO-tr using random sub-sampling and tested on the GEO-te data.
The tables below indicate the MAE and standard deviation between subsamples, obtained by each model.
\subsection{Performance}
Our best performing model uses the Lipschitz Adaptive Learning Rate along with the A-ReLU activation function, and is trained for $200$ epochs. The results are summarized in Tables I,II,III and IV. For the A-Relu activation function, we use $k = 0.6$ and $n = 1.2$. The learning rate in the fixed LR experiments is set to $5 \times 10^{-7}$. The starting value for the Lipschitz Adaptive LR depended on the random initialization of the network weights.

\begin{table}[h!]
\caption{MAE Comparison: Our Model trained on Google Colab Vs D-GEX: Both models are trained on entire corpus of training data. Our model does better than D-GEX with constrained infrastructure (One hidden layer less)}
\label{table:1}

\begin{center}\begin{tabular}{|p{3.2cm}||p{2.0cm}|}
\hline
\multicolumn{2}{|c|}{Training on the Entire data set} \\
\hline
 Model & MAE \\
 \hline
 D-GEX(D) &\\  3-layer architecture &\\ {$9000 \times 3$} &0.3240 \\
 \hline
 D-GEX(L)(A-ReLU) &\\ 2-layer architecture &\\ {$9000 \times 2$} &0.3213\\
\hline
\end{tabular}\end{center}
\end{table}

\begin{table}[h!]
\caption{Comparison of the Smallest D-GEX(D) architecture Trained on Entire Dataset and D-GEX(L) Trained on Random subsamples(Less Than 1/5 Size: D-GEX(D) performs marginally better but held the advantage of training over the entire data set.)}
\label{table:1}

\begin{tabular}{ |p{1.8cm}||p{1.8cm}|p{1.8cm}|p{1.8cm}|}
 \hline
 \multicolumn{4}{|c|}{} \\
 \hline
 Epochs &D-GEX(D) (3000) &D-GEX(L) (500) &D-GEX(L) (1500)\\
 \hline
 100 &-- &0.380709  &0.367214\\
 200 &0.3421   &0.378030  &0.364621\\
      
 \hline
\end{tabular}
\end{table}

\begin{table}[h!]
\caption{Comparison Of MAE on D-GEX(L) Vs D-GEX(D): Both models are trained on identical subsamples: Our model D-GEX(L) performs better. }
\label{table:1}

\begin{tabular}{|p{1.15cm}||p{1.15cm}|p{1.15cm}|p{1.15cm}||p{1.15cm}|}
\hline
\multicolumn{5}{|c|}{Training on sub-sampled datasets} \\
\hline

 Epochs &D-GEX(D) (500) &D-GEX(D) (1500) &D-GEX(L) (500)& D-GEX(L) (1500) \\
 
 \hline
 100 & 0.421837 & 0.399483 & 0.380709 & \textbf{0.367214} \\
 200 & 0.388720 & 0.396381 & 0.378030 & \textbf{0.364621} \\
      
 \hline
 \multicolumn{5}{|c|}{0.0001 $\leq$ $\sigma$ $\leq$ 0.0007} \\
 \hline
\end{tabular}
\end{table}

\begin{table}[h!]
\caption{Mean Absolute Error when training D-GEX(D): Trained on Subsamples}
\label{table:1}

\begin{tabular}{ |p{1.05cm}||p{1.05cm}|p{1.05cm}|p{1.05cm}|p{1.05cm}|p{1.05cm}|}
 \hline
 \multicolumn{6}{|c|}{MAE after training with D-GEX(D) with different choices of activations} \\
 \hline
 Number Of Neurons &Number of Epochs &Sigmoid & Tanh & ReLU &A-ReLU\\
 \hline
 500  &200 &0.438009   &0.396381 &\textbf{0.390081}  &0.394212\\
 1500 &200 &  0.389272  &0.388720 &0.381064  &0.378030\\
      
 \hline
 \multicolumn{6}{|c|}{0.00008 $ \leq $ $\sigma$ $ \leq $ 0.0003} \\
 \hline
\end{tabular}
\end{table}

\begin{table}[h!]
\caption{Mean Absolute Error when training D-GEX(L): Trained on Subsamples}
\label{table:1}

\begin{tabular}{ |p{1.05cm}||p{1.05cm}|p{1.05cm}|p{1.05cm}|p{1.05cm}|p{1.05cm}|}
 \hline
 \multicolumn{6}{|c|}{MAE after training D-GEX(L) with different choices of activations} \\
 \hline
 Number Of Neurons &Number of Epochs &Sigmoid & Tanh & ReLU &A-ReLU\\
 \hline
 500  & 200 &  0.389381  &0.388309 & 0.381064   &0.378030 \\
 1500 &200 &0.374043   &0.376105& 0.368724  &\textbf{0.364621}\\
      
 \hline
 \multicolumn{6}{|c|}{0.0001 $\leq$ $\sigma$ $\leq$ 0.0006} \\
 \hline
\end{tabular}
\end{table}

\begin{table}[h!]
\caption{Mean Absolute Error when training D-GEX(F):Trained on Subsamples}
\label{table:2}

\begin{tabular}{ |p{1.05cm}||p{1.05cm}|p{1.05cm}|p{1.05cm}|p{1.05cm}|p{1.05cm}|}
 \hline
 \multicolumn{6}{|c|}{MAE after training with D-GEX(F) with different choices of activations} \\
 \hline
 Number Of Neurons &Number of Epochs &Sigmoid & Tanh & ReLU &A-ReLU\\
 \hline
 500   & 200 &0.422051    &0.395568 & 0.387693 &0.387512\\
       
 1500 & 200 &  0.402797  &0.380132 &0.370924 &0.370296\\
 \hline
 \multicolumn{6}{|c|}{0.00004 $\leq$ $\sigma$ $\leq$ 0.0004} \\
 \hline
\end{tabular}
\end{table}

\label{sec:others}
As expected, reflecting the base paper, the larger among the two architectures used performed the best, as larger architectures generally allow for a richer representation of features.
\subsubsection{The effect of Lipschitz adaptive learning rate}
It is apparent that in all configurations, architectures employing the use of the LALR, (D-GEX(L)) converge better than those using a fixed learning rate. Using a Lipschitz adaptive scheduler allows the learning rate to be appropriately large away from the minima, allowing it to converge faster, and adequately small so as to not overshoot the minima when in its vicinity. 

As seen in Figure 1, the learning rate while using the scheduler showcases a seemingly exponential reduction while training, and is significantly larger than the learning rate obtained while using the Decay Factor based scheduler used by the authors in \cite{chen2016gene}, while not being too large. This allows a much faster convergence.

Given the difficulty of the problem, the Lipschitz adaptive learning rate allows much faster convergence and makes it possible to carry out training even with minimal computing infrastructure with a reasonable training time.

\begin{figure}[!t]
    \centering
        \centering
        \includegraphics[width=2.5in]{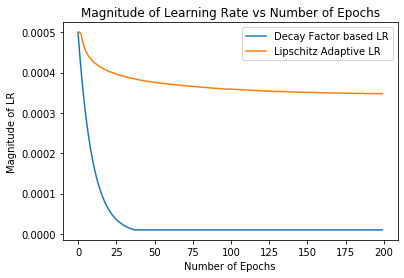}
        \caption{Lipschitz Adaptive Learning Rate is always significantly higher, than the Decay Based Learning Rate employed in \cite{chen2016gene} allowing much faster convergence}
        \label{fig:fig2}
\end{figure}

\begin{figure}[!t]
    \centering
        \centering
        \includegraphics[width=2.5in]{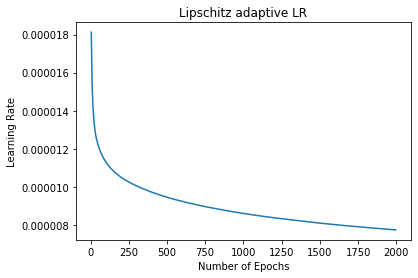}
        \caption{Learning Rate using Lipschitz adaptive LR over 2000 epochs. The learning rate showcases an exponential decrease}
        \label{fig:fig2}
\end{figure}
\subsubsection{The effect of using A-ReLU}
Another interesting observation is the success of A-ReLU as an activation function when training in conjunction with Lipschitz scheduler. The original D-GEX paper trains the network for 200 epochs using the Tanh activation function. Comparing the performance of standard activation functions, and for the same number of epochs, we observe that A-ReLU performs the best.

\begin{figure}[!t]
    \centering
        \centering
        \includegraphics[width=2.5in]{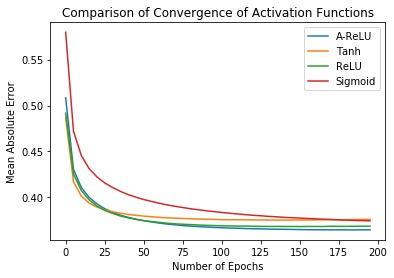}
        \caption{MAE from 0 to 200 epochs}
        \label{fig:fig2}
    \vfill
        \centering
        \includegraphics[width=2.5in]{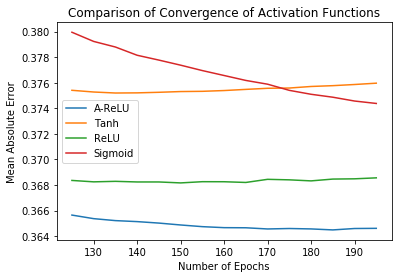}
        \caption{MAE from 125 to 200 epochs \\ \\ The two plots above indicate how the A-ReLU activation function allows the network to converge to smaller errors than other, more conventional activation functions in this task}
        \label{fig:fig3}
\end{figure}

\section{Discussion and Conclusion}
We demonstrate that neural network learning of computational biology tasks like gene expression inference from huge Genomic data sets can be achieved with limited computational resources, specifically only one GPU with 4 GB RAM. We also present evidence that a combination of a shallow neural network and an adaptive learning rate achieve better performance than deeper nets.

We divide the GEO dataset into smaller random sub-samples in order to handle the massiveness of the training data. To ensure that no bias due to the exclusion of samples because of the random nature of sub-sampling, the reported results are obtained by averaging the MAEs corresponding to each one of six sub-samples. In order to accelerate training, an adaptive learning rate based on the Lipschitz constant has been utilized. The performance is also subjected to different activation functions to capture significant deviation, if any. A-ReLU is found to outperform several other, better known activation functions.

We obtain similar results to the original D-GEX architecture \cite{chen2016gene} with a neural network containing a single hidden layer with 1500 neurons and trained on data, 1/5th the size of the original data set. This is reported in Table II. However, when the original D-GEX architecture and model was employed on smaller subsets of data, it is observed that our methods perform better (please see Table III). In order to establish the merit of our contribution further, we ran a final set of experiments on the entire training corpus, on Google COLAB, with a Tesla K80 GPU having 12 GB RAM. We show, in Table I, that we have again accomplished better performance in comparison with the base paper\cite{chen2016gene}. Apart from one,all the architectures used in this work consist of a single hidden layer, with a maximum of 1500 neurons. The largest architecture utilized in \cite{chen2016gene} was composed of three hidden layers of 9000 neurons each. As opposed to NVIDIA TITAN Z GPUs used in the base paper\cite{chen2016gene}, all training in this work was done using an NVIDIA GeForce 1050Ti GPU possessing upto 4 GB RAM, on a laptop, except for one case which shows that our methods can be used at-scale.
\par The learning rate computed and used in training in this paper is truly adaptive and the term is not loosely used.  The new learning rate at every iteration is computed by the formula presented in section IV. This is done without manual intervention and any sort of camouflaged supervision. Such Lipschitz adaptive learning rate is, of course, dependent on the choice of loss function and will vary based on the loss used in training and provided that the loss function is continuous and smooth up to the first order i.e. Lipschitz. The remarkable outcome of developing a LALR based training is evident in larger architectures as well; In Table I we demonstrate that, our network of 2 hidden layers of 9000 neurons  beats the performance of a deeper neural net with 3 hidden layers of 9000 neurons each, hence performing better than a network with an entire additional layer. Furthermore, the best results produced in \cite{chen2016gene} uses a total of $27,000$ neurons whereas we have made use of only $18000$ neurons in one experiment, and a maximum of $1500$ neurons in all others. Our results are embellished with multiple runs on the data set with demonstrated insignificant standard deviation between MAEs computed for each run. Thus, we establish our claim of accomplishing impressive results via Parsimonious Computing. \footnote{Our code can be found at: \emph{\href{https://github.com/ShaileshSridhar2403/Parsimony-Computing-Gene-Expression-Inference-with-LALR}{https://github.com/ShaileshSridhar2403/Parsimony-Computing-Gene-Expression-Inference-with-LALR}}}
\section*{Acknowledgement}
The authors would like to thank the Science and Engineering Research Board (SERB)-(DST), Government of of India for supporting this research (SERB-EMR/ 2016/005687). 
\nocite{*}
\bibliography{references} 
\bibliographystyle{unsrt}

\end{document}